\documentclass{article}
\usepackage{spconf,amsmath,graphicx}

\usepackage{microtype} 
\usepackage{times}
\usepackage{url}
\usepackage{latexsym} 
\usepackage{float} 
\usepackage{graphicx}
\usepackage{amsmath}
\usepackage{amsthm}
\usepackage{booktabs}
\usepackage{algorithm}
\usepackage{algorithmic}
\usepackage{stfloats}
\usepackage{multirow}
\usepackage{color}
\usepackage{subfigure} 
\usepackage{latexsym}
\usepackage{makecell}

\usepackage{enumitem}

\title{Joint intent detection and slot filling based on continual learning  model}
\name{Yanfei Hui, Jianzong Wang$^\ast$ \thanks{*Corresponding author: Jianzong Wang, jzwang@188.com}, Ning Cheng, Fengying Yu, Tianbo Wu, Jing Xiao}
\address{Ping An Technology (Shenzhen) Co., Ltd.}
\begin{document}

\maketitle

\begin{abstract}
Slot filling and intent detection have become a significant theme in the field of natural language understanding. 
Even though slot filling is intensively associated with intent detection, the characteristics of the information required for both tasks are different while most of those approaches may not fully aware of this problem. 
In addition, balancing the accuracy of two tasks effectively is an inevitable problem for the joint learning model.
In this paper, a Continual Learning Interrelated Model (\textbf{CLIM}) is proposed to consider semantic information with different characteristics and balance the accuracy between intent detection and slot filling effectively. The experimental results show that CLIM achieves state-of-the-art performace on slot filling and intent detection on ATIS and Snips. 
\end{abstract}


\section{Introduction}
Natural Language Understanding (NLU) typically includes the intent detection and slot filling tasks, aiming to identify intent and extract semantic constituents from the utterance.
Intent detection is essentially a classification problem and therefore popular classifiers like Linear Regression (LR), K-Nearest Neighbor (KNN), Support Vector Machines (SVMs) and deep neural network methods can be applied \cite{Sarikaya2011DeepBN,Lai2015RecurrentCN}. 
Slot filling should be considered as a sequence labeling task that tags the input word sequence \(\boldsymbol x=(\boldsymbol x_1,\boldsymbol x_2,...,\boldsymbol x_T)\) with the slot label sequence \(\boldsymbol y=(y_1,y_2,...,y_T)\). 
It should be noted that alignment is explicit in slot filling unlike machine translation and speech recognition, thus alignment-based models typically work well\cite{Sutskever2014SequenceTS,Chan2016ListenAA}.
Popular approaches to solving sequence labeling problems include Conditional Random Fields (CRFs), and Recurrent Neural Networks (RNNs) based approaches \cite{Mikolov2011ExtensionsOR}. 
Particularly Gated Recurrent Unit (GRU) and Long Short-Term Memory (LSTM) models have achieved good performance.

Due to the accumulation of errors, pipeline methods usually fail to achieve satisfactory performance. 
Accordingly, some works suggested using one joint model for slot filling and intent detection to improve the performance via mutual enhancement between two tasks. 
Recently, several joint learning methods for intent detection and slot filling were proposed to improve the performance over independent models \cite{Guo2014JointSU,Hakkani2016MultiDomainJS}. 
It has been proved that attention mechanism \cite{Bahdanau2014NeuralMT} can improve model's performance by dealing with long-range dependencies than currently used RNNs. 
So far attention-based joint learning methods have achieved the state-of-the-art performance for joint intent detection and slot filling \cite{Liu2016AttentionBasedRN,Chen2019BERTFJ}. 
 
However, the prior work seems to lose sight of the fact that slot filling and intent detection are strongly correlative and the two tasks need encoding information with different characteristics.
Among them, slot filling task needs more accurate encoding information.
And there is a problem of alignment between slot and word, while intent detection only focuses on the overall semantic information, which is not sensitive to the word location information.
Apart from that, there is a phenomenon of "seesaw" in the later stage of training the joint model, that is to say, when the accuracy of one task increases, it often causes the accuracy of the other task to decline.  

\begin{figure}[h]
\setlength{\abovecaptionskip}{0.cm}
\setlength{\belowcaptionskip}{-0.cm}
\centering
\includegraphics[scale=0.65]{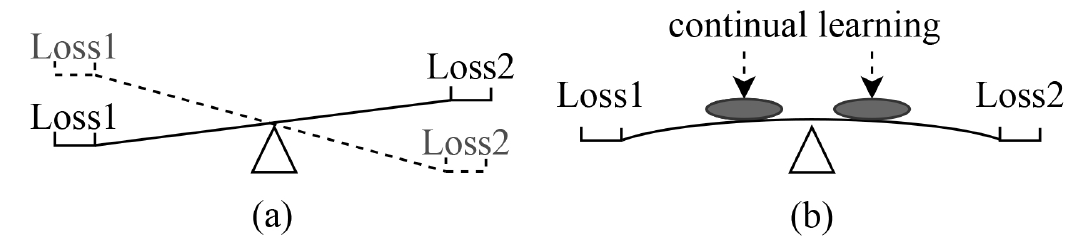}
\small
\caption{Multitask joint training process}
\label{Figure1}
\end{figure}


As shown in Figure 1a, when using the joint model to solve two tasks at the same time, the increase of task 1 accuracy will lead to the deterioration of task 2 performance, and vice versa. 
In this paper, CLIM is proposed to further enrich encoding information and take the cross-impact between two task into account.  
We introduce the idea of continuous learning into the training process, which makes the model perform satisfactorily in both tasks, as shown in Figure 1b.

In summary, the contributions of this paper are as follows:

\setlist{nolistsep}
\begin{itemize}

\item Applying the idea of continuous learning to solve the phenomenon of “precision seesaw” in the process of multi task training.
\item Proposing a variant dual-encoder model structure to enrich the semantic information and location information contained by adopting different encoding forms.
\end{itemize}

\section{Proposed Approach} 

\subsection{Variant Dual-encoder Interrelated Model} 
We make a novel attempt to bridge the gap between word-level slot model and the sentence-level intent model via a dual-encoder model architecture, as shown in Figure 2.  

\begin{figure}[!h]
\centering
\includegraphics[scale=0.48]{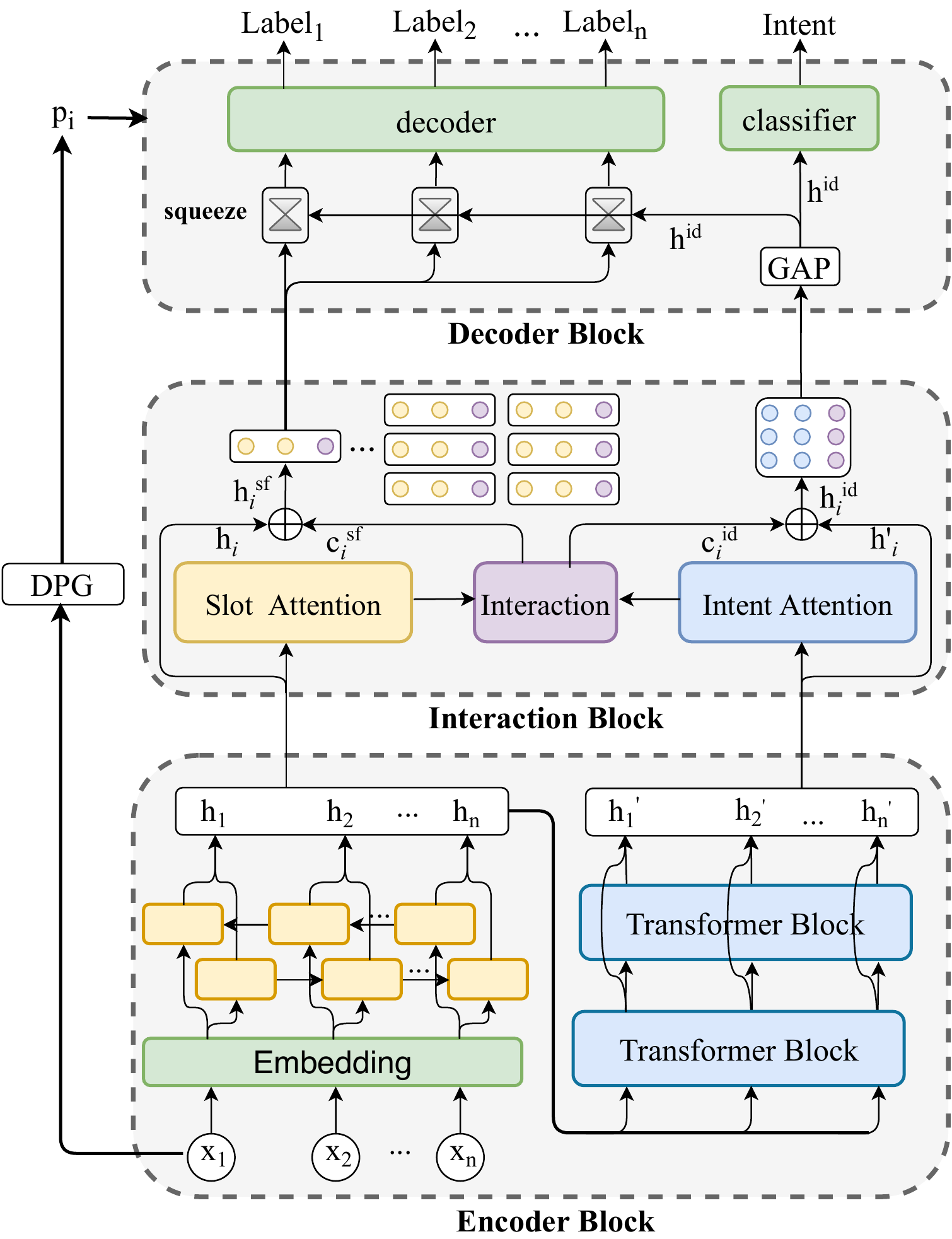}
\small
\caption{Variant Dual-encoder Interrelated Model}
\label{Figure2}
\end{figure}
 
\subsubsection{Variant Dual-encoder}
As mentioned above, CLIM apply two encoders (RNN and Transformer) to encode the sentence separately. 
As shown in Figure 2, there are two encoders in the Encoder Block, one on the left is a Bi-directional Long Short-Term Memory (BiLSTM), which can encode the input information in sequence. 
The additional encoder on the right is a bilayer transformer block \cite{Vaswani2017AttentionIA}.
Transformer models constitute the current state-of-the-art in machine translation and are becoming increasingly popular also in other domains \cite{Li2018NeuralSS,Devlin2018BERTPO,john2020a}, such as language modelling.  
 
The BiLSTM reads in the input sentences \((\boldsymbol x_1,\boldsymbol x_2,...,\boldsymbol x_n)\) forward and backward, and generates two sequences of hidden states \(\boldsymbol h^f_t\) and \(\boldsymbol h^b_t\). 
A concatenation of \(\boldsymbol h^f_t\) and \(\boldsymbol h^b_t\) forms a final BiLSTM state \(\boldsymbol h_t=concat [\boldsymbol h^f_t, \boldsymbol h^b_t]\) at time step \(t\). 
On the other side, the hidden states generated by two transformer blocks are joined together by residual mechanism. 
In order to make the model have a faster inference speed, CLIM directly takes the output of LSTM as the input of the double-layer transformer block, so as to get another different granularity encoding information. Then this encoder can also get a hidden state sequence \((\boldsymbol h'_1,\boldsymbol h'_2,...,\boldsymbol h'_n)\) : \(\boldsymbol h'_i =tanh(\boldsymbol h^1_i+\boldsymbol h^2_i)\), where the hidden state \(\boldsymbol h^1_i\) stand for the output of first transformer block, and the hidden state \(\boldsymbol h^2_i \) represents the output of the second transformer block at time step \(i\).
The \(i^{th}\) context vector \(\boldsymbol c^{sf}_i\) in the slot attention matrix and \(\boldsymbol c^{id}_i\) in the intent attention matrix are computed as the weighted sum of encoders' hidden states:
\(\boldsymbol c^{sf}_i=\sum\limits_{j=1}^T{\boldsymbol \alpha}_{i,j}\boldsymbol h_j\) and
\(\boldsymbol c^{id}_i=\sum\limits_{j=1}^T{\boldsymbol \alpha'}_{i,j}\boldsymbol h'_j\).
The attention weight \(\boldsymbol \alpha\) is acquired the same as that mentioned introduced in \cite{Liu2016AttentionBasedRN}.  
And the superscript 'sf' and 'id' represent slot filling and intent detection respectively.
More formally, let \(\boldsymbol h_j,\boldsymbol h'_j \in \boldsymbol R^{d\times s}\) be the \(j^{th}\) sentence hidden states obtained by two encoders, respectively.

\subsubsection{Interaction Modes between Encoders}
Then, in the Interaction Block of CLIM, two attention matrices are used for information interaction by using this mechanism: \(\boldsymbol c^{sf}_i , \boldsymbol c^{id}_i= \boldsymbol c^{id}_i , \boldsymbol c^{sf}_i\).
The interaction mechanism can be defined in a variety of ways. We found that a straightforward implementation using matrix exchange can already achieve good results. In the Interaction Block, two kinds of hidden state vectors with different granularity are joined together and then sent to the Decoder Block to complete slot filling and intent detection tasks.
The hidden state vector \(h^{sf}\) and \(h^{id}\) that is eventually fed to the Decoder Block can be obtained by concate function:  \(\boldsymbol h_i^{sf}=concat[\boldsymbol h_i,\boldsymbol c_i^{sf}]\) and \( \boldsymbol h_i^{id}= concat[\boldsymbol h'_i,\boldsymbol c_i^{id}]\).
Then, the final hidden state used for intent detection can be obtained by global average pooling.
As shown in Figure 2, the Decoding Layer includes a decoder and a classifier for slot filling and intent detection, respectively. Next we introduce the details of decoder and classifier.
 
\subsubsection{Decoding Layer }

For the classification task of intent detection, CLIM obtains the final intent category distribution by fusing hidden state  \(\boldsymbol h_i^{id}\). The classifier obtains the final intent in the following ways:
\begin{equation}
y_{intent}=softmax(W_{intent}\cdot GAP(\boldsymbol h_i^{id}))  
\end{equation} 
Besides, CLIM compressed the input information of decoder which can remove the information noise only related to the training set, so as to improve the generalization ability of the model. 
The compression process is realized by using one-dimensional convolution operation. 
\begin{equation}
\boldsymbol s_i=conv1d(concat[\boldsymbol h_i^{sf},\boldsymbol h^{id}])  
\end{equation}
where \(\boldsymbol s_i\) is the input for decoder of slot filling. At last, the compressed dense representation will be used for sequence annotation.

 

\subsection{Dynamic Parameter Generation} 
Several neural networks can be used to implement Dynamic Parameter Generation (DPG) for parameter generation, e.g., convolutional neural network (CNN), RNN and multilayer perceptron (MLP). 
The objective of this paper is not to explore all possible implementations. 
We found that a straightforward implementation using MLP can already achieve good results. Formally, DPG can be written as:
\begin{equation}
\boldsymbol p_i = f(\boldsymbol z_i; \mu) = \sigma(\boldsymbol w_D\boldsymbol z_i + \boldsymbol b)
\end{equation}
where \(\mu\) is the activation function, and \(\boldsymbol w_D\) and b are the parameters of DPG and denoted by \(\mu\).

\section{Experiments and Results}
\subsection{Data}
ATIS (Airline Travel Information Systems) data set contains audio recordings of people making flight reservations. 
As described in Table 2, ATIS data set contains 4978 train and 893 test spoken utterances. 
We also use Snips, which is collected from the Snips personal voice assistant. This data set contains 13084 train and 700 test utterances. We evaluate the model performance on slot filling using F1 score, and the performance on intent detection using classification error rate.
 
\begin{table}[h]
\setlength{\abovecaptionskip}{0.cm}
\setlength{\belowcaptionskip}{-0.cm}
\caption{Dataset details statistics} 
\small
\begin{center}
\begin{tabular}{p{3.6cm}|p{1.4cm}|p{1.4cm} } 
\hline
Dataset &ATIS  &Snips\\
\hline
Vocab Size& 722&11241 \\
Average Sentence Length&11.28 &9.05\\
Intents &21 &7\\
Slots&120&72 \\
Training Samples&4478&13084 \\
Validation Samples&500 &700\\
Test Samples&893&700 \\
\hline
\end{tabular}
\end{center}
\end{table}
 
\begin{table*}[b]
\setlength{\abovecaptionskip}{0.cm}
\setlength{\belowcaptionskip}{-0.cm}
\caption{Performance of different joint models on two datasets.}
\centering
\smallskip
\smallskip
\footnotesize
\begin{tabular}{p{7.5cm}<{\raggedright}|p{1.5cm}<{\centering} p{1.5cm}<{\centering}|p{1.5cm}<{\centering} p{1.5cm}<{\centering}}
\hline
\multirow{2}{*}{Model}  & \multicolumn{2}{c|}{ATIS} & \multicolumn{2}{c}{SNIPS} \\
& Slot(F1)&Intent(Acc) &Slot(F1)&Intent(Acc)\\ 
\hline 
Atten. enc-dec NN with aligned inputs\cite{Liu2016AttentionBasedRN} &95.87&98.43&-&-\\
Atten.-BiRNN\cite{Liu2016AttentionBasedRN}  &95.98&98.21&-&- \\
Enc-dec focus\cite{Su2017ENCODER}&95.79&-&-&- \\
Slot-Gated\cite{Goo2018SlotGatedMF}&95.2& 94.1&88.8&97.0\\
CAPSULE-NLU\cite{zhang2019joint}  &95.2&95.0&91.8&97.3\\
SF-ID network(with CRF)\cite{Haihong2019ANB}&95.75&97.76&91.43&97.43\\
Joint BERT\cite{Chen2019BERTFJ}&96.1&97.5&97.0&98.6\\
\hline
\hline
CLIM&\textbf{96.34}&\textbf{98.43}&\textbf{97.03}&\textbf{99.00}\\
\hline
\end{tabular}
\end{table*}

\subsection{Training Procedure}
LSTM cell is used as the basic RNN unit in the experiments.
Given the size the data set, our proposed model set the number of units in LSTM cell as 200.  
Word embeddings of size 1024 are pre-trained and fine-tuned during mini-batch training with batch size of 20. 
Dropout rate 0.5 is applied behind the word embedding layer and between the fully connected layers. 
And our proposed model use Adam optimization method. In the process of model training, after a certain number of epochs, we tried to fix the weight of the word embedding layer so that the parameters would not be updated.

\subsection{Joint Learning}
Table 2 shows our joint training model performance on intent detection and slot filling comparing to previous reported results where the compared baselines include the state-of-the-art sequence-based joint model using bidirectional LSTM  \cite{Hakkani2016MultiDomainJS} and slot-gated model \cite{Goo2018SlotGatedMF}. 
As shown in this table, our proposed model CLIM achieves \textbf{96.34$\%$} F1 score on slot filling task and \textbf{98.43$\%$} intent classification accuracy on ATIS dataset. 
Comparing with other models, it can be found that CLIM has the same accuracy as the optimal result in intent detection task and achieves state-of-the-art on slot filling task, even better than BERT, the pre-trained model with large parameters \cite{Chen2019BERTFJ}.
Moreover, on the Snips dataset, it can be found that although F1 score only slightly improves on slot filling tasks, our proposed model has made significant progress in the accuracy of intent detection. 
CLIM reduces the error rate of intent detection from \textbf{1.4$\%$} to \textbf{0.84$\%$} , and achieves \textbf{40.0$\%$} reduction in error rate.

\subsection{Continual Learning}
It is found that if the traditional continuous learning method is used to train the model, the intent detection task will be learned first, and then the slot filling task will be learned, or vice versa \cite{Hu2019OvercomingCF}. 
However, through the previous experiments, we have found that the performance of solving two tasks at the same time by using the joint model is more than training on one task alone, so we need to modify the continuous learning method.
Figure 3 and Figure 4 show the independent/joint/continual learning performance on intent detection and slot filling.
We find CLIM loses 0.06$\%$ for slot filling on ATIS
but wins joint learning 0.05$\%$ on Snips. 
We can observe that introducing more parameters does not always perform better. 
The different performance of continuous learning in two datasets should be due to the different size of vocabulary and types of slots.
Meanwhile, CLIM wins joint learning 0.16$\%$ for intent detection on Snips and achieves the same result on ATIS. 
It makes sense since the classifier could use features
from LSTM and Transformer simultaneously and different
model structures complement each other for classification. 
In particular, our architecture performs best for both cases. 
In fact, DPG does not introduce new model structure or complicated operations and the number of parameters are almost the
same. 
 
\subsection{Contrast Study}
To further evaluate the advances of variant dual-encoder architecture for joint learning, we tested different dual-encoder structures. 
Note that all the variants are based on joint learning with information compression mechanism :
\begin{itemize}
\setlength{\itemsep}{0pt}
\setlength{\parsep}{0pt}
\setlength{\parskip}{0pt}
\item BiLSTM-BiLSTM[B-B], where both encoders are bidirectional LSTM.
\item BiLSTM-Transformer[B-T], where the left encoder is a bi-directional LSTM, and the other encoder is a double-layer transformer with residual connection.
\item BiLSTM-Transformer(V)[B-T(V)], where the input of transformer is no longer word vector, but the encoding result of bidirectional LSTM. This is the structure adopted by CLIM model in this paper.
\end{itemize}

\begin{figure*} 
\begin{minipage}[t]{0.5\linewidth} 
\centering 
\includegraphics[width=2.5in]{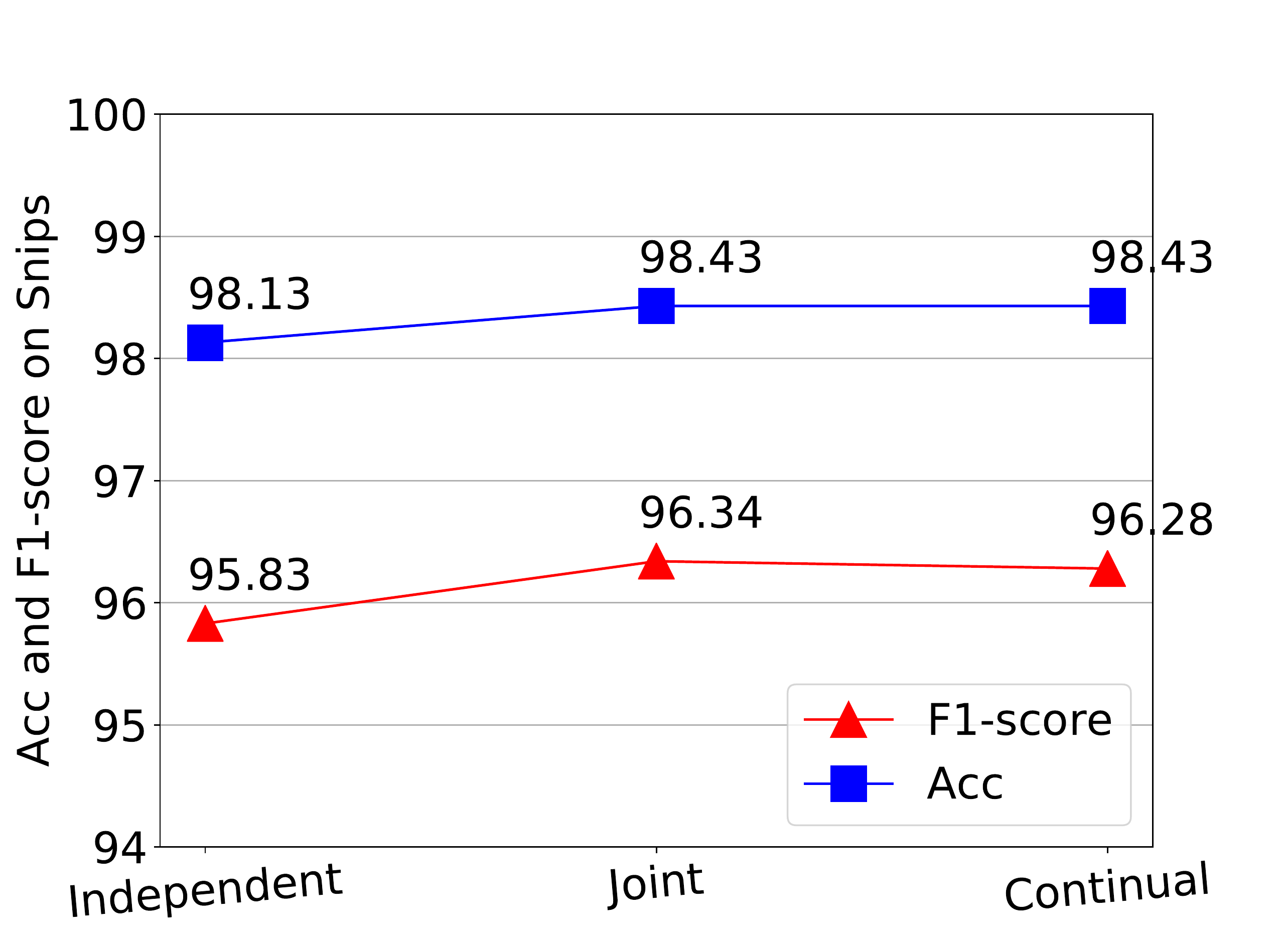} 
\caption{Performance of three methods on ATIS } 
\label{frame} 
\end{minipage}%
\begin{minipage}[t]{0.5\linewidth} 
\centering 
\includegraphics[width=2.5in]{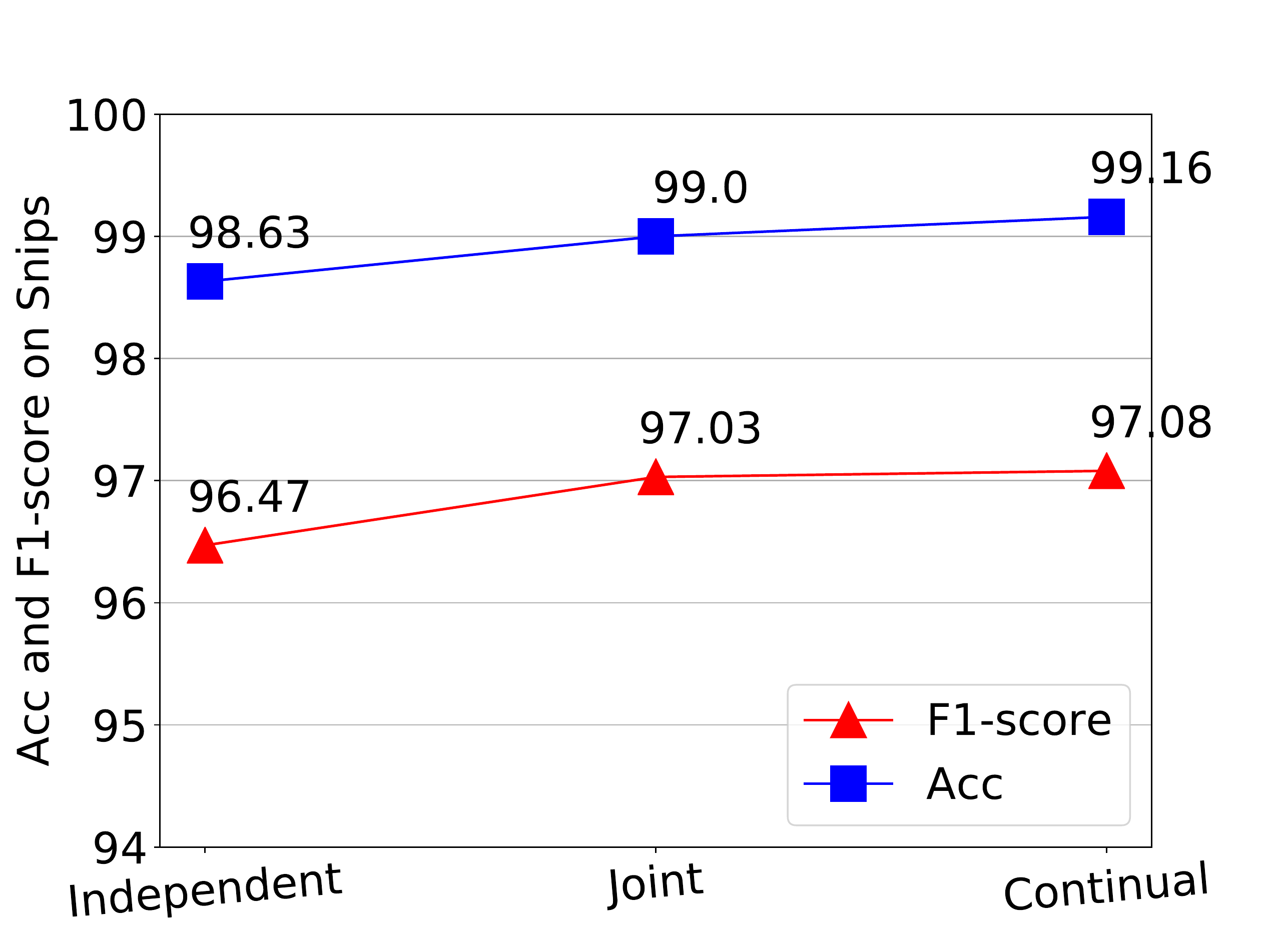} 
\caption{Performance of three methods on Snips } 
\label{label} 
\end{minipage} 
\end{figure*}
 
 \begin{figure}[htbp]
 \setlength{\abovecaptionskip}{0.cm}
\setlength{\belowcaptionskip}{-0.cm}
\centering
\includegraphics[scale=0.65]{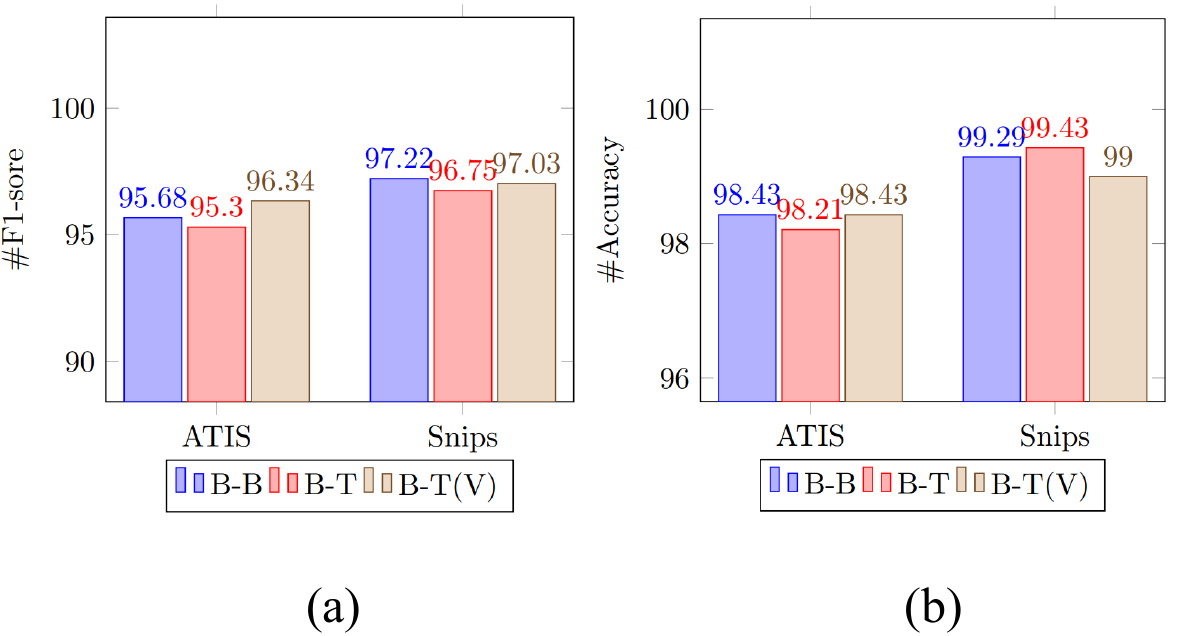}
\small
\caption{Performance of different encoders on two datasets.}
\label{fig:encoder_decoder0831}
\end{figure}
Figure 5 shows the joint learning performance of our model on ATIS data set and Snips data set by changing the structure of dual-encoder at a time. We found that the F1 score of slot filling on ATIS dataset decreased slightly with BiLSTM-BiLSTM, but the model can achieve the optimal accuracy in intent detection task. More importantly, it can achieve state-of-the-art for two tasks on Snips dataset at the same time. BiLSTM-Transformer does not perform well on ATIS dataset, and although it improves the accuracy of slot filling task in Snips dataset, the F1 score in slot filling task decreases a lot.
Due to the small number of stack layers of transformer, it is difficult to play the encoding ability as powerful as BERT. 
BiLSTM-Transformer(V) can achieve state-of-the-art for two tasks on ATIS dataset at the same time. Although CLIM model has better results than other models on Snips dataset, the improvement of F1 score and accuracy is not as good as BiLSTM-BiLSTM. 
As shown in Figure 5, the model with BiLSTM-BiLSTM performs better than CLIM on Snips dataset, which is due to shorter sentences of Snips compared with ATIS dataset. 
As listed in the table, the variant dual-encoder model struture contributes to both slot filling and intent classification task.

 
\section{Conclusion}
In this paper, we proposed a continual learning model architecture to address the problem of accuracy imbalance in multitask. 
Using a joint model for the two NLU tasks simplifies the training process, as only one model needs to be trained and deployed.
Our proposed model achieves state-of-the-art performance on the benchmark ATIS and SNIPS. 
\section{Acknowledgements}
This paper is supported by National Key Research and Development Program of China under grant No. 2017YFB1401202, No. 2018YFB1003500, and No. 2018YFB0204400. Corresponding author is Jianzong Wang from Ping An Technology (Shenzhen) Co., Ltd.

\vfill\pagebreak
  
\bibliographystyle{IEEEbib}
\bibliography{strings,refs,mybib}

\end{document}